\pdfoutput=1
\documentclass[11pt]{article}

\usepackage{EMNLP2022}

\usepackage{times}
\usepackage{latexsym}

\usepackage[T1]{fontenc}
\usepackage[utf8]{inputenc}

\usepackage{microtype}

\usepackage{inconsolata}

\usepackage{amsmath}
\usepackage{graphicx}
\usepackage{booktabs}

\title{HYRR: Hybrid Infused Reranking for Passage Retrieval}

\author{
    Jing Lu, Keith Hall, Ji Ma \and Jianmo Ni \\
    Google Research \\
    \texttt{\{ljwinnie,kbhall,maji,jianmon\}}@google.com \\
}

\begin{document}
\maketitle
\begin{abstract}
We present Hybrid Infused Reranking for Passages Retrieval (\textbf{HYRR}), a framework for training rerankers based on a hybrid of BM25 and neural retrieval models. Retrievers based on hybrid models have been shown to outperform both BM25 and neural models alone. Our approach exploits this improved performance when training a reranker, leading to a robust reranking model. The reranker, a cross-attention neural model, is shown to be robust to different first-stage retrieval systems, achieving better performance than rerankers simply trained upon the first-stage retrievers in the multi-stage systems. We present evaluations on a supervised passage retrieval task using MS MARCO and zero-shot retrieval tasks using BEIR. The empirical results show strong performance on both evaluations.

%show
%improves over supervised baselines and SOTA zero-shot approaches on 11 datasets on BEIR when we adapt the reranker to the target domain using a synthetic question generation adaptation approach.
\end{abstract}

\section{Introduction}
Information Retrieval (and subsequent retrieve-and-read approaches to open-domain question answering (QA)) have generally seen pipelined retrieve-and-rank approaches achieve the best accuracies.  This paradigm has become so common that many of the TREC IR evaluations have placed an emphasis on reranking the top N (often 1000) results from a first-pass retriever \cite{trecdl2021,trec-fair2021}. There has been significant work recently showing that neural retrieval models can improve recall and produce state-of-the-art retrievers \cite{dpr,rocketqa,rocketqav2}, but this is not true across the board; there are still many tasks where simple term-based techniques such as BM25 \cite{bm25} significantly outperform sophisticated neural retrieval models. %\jing{find a new example}
%For example, Karpukhin et al. \shortcite{dpr} showed that BM25 beats dense retrieval models on SQuAD \cite{squad}.

%For example, Ma et al.\shortcite{bioasq2020} showed that BM25 beats dense retrieval models on BioASQ challenge \cite{bioasq-overview}. %\textit{TODO: Add specific pointers to tasks where BM25 is still superior and to recent publications showing DPR-style is superior for NQ.} TODO: find another example

This motivates recent exploration in hybrid retrievers in order to join the best from each world. Techniques which combine term-based models and neural models in either early stage \cite{QGen} or later stage \cite{dpr,ecir22} outperform each individual model. No matter how much improvement achieved from first-stage retrievers, the reranking model continues to achieve an additional performance boost. 

In this work, we revisit the question of how best to train a reranking model for retrieval. We study two different settings: the supervised setting where we have training data in the target domain, and the zero-shot setting where we have no training data in the target domain. The traditional wisdom is that you should train the reranker on data that is similar to the distribution that you will observe at inference time, and the best means to do so is to use the same retriever for inference as you do for training the reranker. That is to say, the first-stage retriever produces candidates for the reranker for both training and inference \cite{multi-stage-bert,reranking-seq2seq}. However, recent work shows that this does not always guarantee an effective reranker \cite{rethink_bert_reranker}.  While we do see this approach achieving strong results in general, we show that by training a robust reranker which has been exposed to training data from a hybrid of term-based and neural retrieval models, we are able to achieve high performance on MS MAMRCO passage ranking task and a large set of out-of-domain datasets (BEIR \cite{beir}) even when reranking candidates are generated by a BM25 retriever.

The primary contributions of this paper are:
\begin{itemize}
\item We present \textbf{HYRR}, a training paradigm for training rerankers based on hybrid term-based and neural retrievers.
\item We show that this approach is effective in both supervised setting and zero-shot setting.
%, achieving an average of 8.8\% improvement when reranking the output of BM25
%, and outperforming other strong baselines by 
\item We show that this approach results in a robust reranker which performs well across different retrievers, domains, and tasks; though there are still limitations which appear to be based in the query generation approach utilized in the zero-shot setting.
\end{itemize}

\section{Related Work}

\subsection{Neural ranking models}
%\begin{itemize}
%    \item bert/t5 ranking
%\end{itemize}

Many prior works explore using neural models for text ranking, most recently the focus as been on transformer-based models \cite{transformer}. Even through it is computationally expensive, a BERT-based cross-attention model is one of the most dominant models for text ranking \cite{bert-reranker, rethink_bert_reranker} because of its capability to model the interaction between the query and passage. Concretely, queries and passages are concatenated and fed into the BERT model, a pairwise score is then obtained by projecting the encoding of [CLS] token. The text ranking problem is cast as a binary classification problem. Recently, encoder-decoder language models, such as T5 \cite{2020t5},  have been adapted for text ranking.  Nogueira et al.\shortcite{t5-reranker} proposed a model that takes a query and passage pair as input of encoder, and the decoder produces the tokens ``true'' or ``false'' to indicate the relevance of a query and a given passage. Pradeep et al. \shortcite{duoT5} further proposed a pairwise ranking model that takes a query and two passages as input and the decoder produces the token ``true'' if the first passage is more relevant then the second passage, and ``false'' otherwise. \newcite{rankT5} proposed T5 encoder-only and encoder-decoder rerankers that optimize ranking performance directly by outputting real-value scores and using ranking losses. \newcite{sgpt} proposed SGPT that uses pre-trained language model as reranking model directly; and \newcite{upr} proposed UPR that uses a zero-shot question generation model via prompting a large language model in order to directly rerank passages.

\subsection{Multi-stage retrieval pipelines}
%\begin{itemize}
%    \item in-pars
%    \item UPR
%    \item GPL (retriever)
%\end{itemize}

Passage retrieval systems commonly adopt the multi-stage pipelined approach where in the first stage, an initial ranking is produced by an efficient model which searches through the entire corpus which may contain millions of passages and in the later stages, the ranking is refined by more precise models. Many existing works use the model in the previous stage to generate training examples for the current stage \cite{bert-reranker,multi-stage-bert,t5-reranker,reranking-seq2seq,inpars}.  Gao et al. \shortcite{rethink_bert_reranker} point out sometimes an effective first-stage model may introduce false positives for the next stages and propose localized contrastive estimation learning.

\section{Model}

Given a query $q_i$ and a list of candidate passages $C(i) = {c_1, c_2, ..., c_n}$ in a document collection $D$, the ranking task aims to sort passages in the $C(i)$ such that more relevant passages have higher scores. More formally, we aim to learn a scoring function $s(q_i, c_j)$ such that ${c^* = argmax_{j\in C(i)} s(q_i, c_j)}$ is the most relevant passage to the query.

\subsection{Model structure}
\label{sec:reranking}

\begin{figure}[t]
    \centering
    \includegraphics[width=0.75\linewidth]{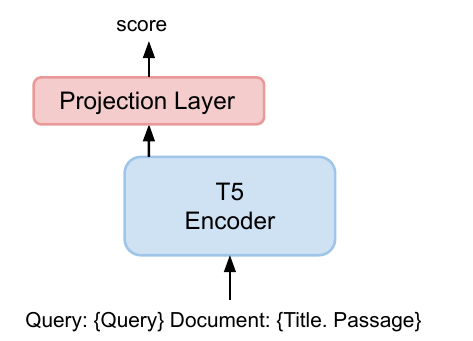}
    \caption{The architecture of T5-based reranking model.}
    \label{fig:reranker}
\end{figure}

We follow \newcite{rankT5} to use a T5-based cross-attention model. Specifically, we represent the query-passage pair as input sequence ``Query: \{Query\} Document: \{Title. Passage\}'' and feed it into the encoder. The output of the encoder is the encodings of the input sequence. We then apply a projection layer on the encoding of the first token and the output is used as the score. We use the encoder and discard the decoder allowing us to exploit the encoder-decoder pretraining while not requiring a decoder for inference. During inference, we pair query $q_i$ with each passage in $C(i)$ and compute scores. The ranking result is obtained by sorting the passages based on their scores. The structure is shown in Figure~\ref{fig:reranker}.

The loss function we use is a listwise softmax cross entropy loss \cite{LTR-softmax} and is defined as follows:

\begin{equation}
% \label{eq:softmax_loss}
    \ell = \sum_{i=1}^{m} \hat{y}_{ij} \log\Big(\frac{e^{s_{ij}}}{\sum_{j'} e^{s_{ij'}}}\Big)   \nonumber
\end{equation}

where $s_{ij}$ is the predicted ranking score on query $q_i$ and passage $c_j$, and $\hat{y}_{ij}$ is the relevance label. 

\subsection{Training data generation}

Usually the number of passages in $C(i)$ is much smaller than the total number of passages in the document collection. How to construct $C(i)$ is important and affects the performance of the ranking model. The widely used strategy is to sample passages returned by a retriever, such as BM25 \cite{sgpt,inpars}. In the pipelined multi-stage retrieval system consisting of retrieval and reranking stages, $C(i)$ is usually formed by using the first-stage retriever. However, a better first-stage retriever not always guarantee a better training set for reranking models. In this work, we present a strategy to train robust rerankers.  

We use a hybrid retriever to generate passage lists, inspired by Ma et al.'s \shortcite{QGen} hybrid first-stage retriever. Specifically, we use a BM25 model as the sparse retrieval model and a T5-based dual encoder model \cite{t5xr} as the dense retrieval model.  For each passage (and query), we concatenate the encodings from the two models to create a hybrid encoding.  We perform maximal inner-product search (MIPS) using approximate nearest neighbor search over these hybrid encodings.  This results in a different set of neighbors than independently selecting neighbors from BM25 and the dual encoder.

For the BM25 model, we represent each query as a $|V|$-dimensional binary encoding $\textbf{q}^{\text{bm25}}$, where $\textbf{q}^{\text{bm25}}[i]$ is $k$ if the i-th entry of vocabulary $V$ is seen $k$ times in the query, 0 otherwise (e.g., this is a vector of query term counts). We represent each passage as a sparse real-valued vector $\textbf{c}^{\text{bm25}}$: 

\begin{small}
\begin{equation}
\textbf{c}^\text{bm25}[i] = 
 \frac{\text{IDF}(t_i)* \text{cnt}(t_i, c)*(k + 1)}{\text{cnt}(t_i, c) + k*(1 - b + b * \frac{m}{m_\text{avg}})} \nonumber
\end{equation}
\end{small}

where $t_i$ are tokens from passage $c$, cnt($t_i$, $c$) is $t_i$'s term frequency in $c$, $k$/$b$ are BM25 hyperparameters, IDF is the term's inverse document frequency from the document collection,  $m$ are the number of tokens in $c$, and $m_\text{avg}$ is the collection's average passage length.  We can recover the original BM25 score via dot-product between the query vector and the passage vector.

\begin{figure}[t]
    \centering
    \includegraphics[width=0.8\linewidth]{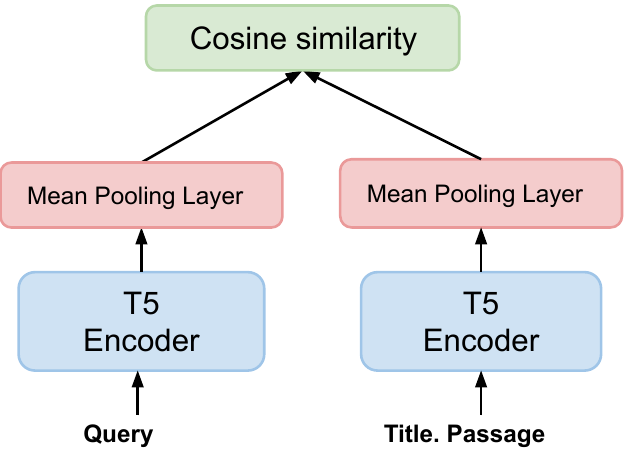}
    \caption{The architecture of T5 dual encoder model.}
    \label{fig:de}
\end{figure}

Our dual encoder is based on T5 \cite{2020t5} as well. We adapt the encoder-only structure from Ni et al.\shortcite{t5xr}'s GTR model. To encode a query, we feed the query text to T5 encoder and use the mean pooling of the encoder output as the query encoding $\textbf{q}^{\text{de}}$. A passage encoding $\textbf{c}^{\text{de}}$ is generated in a similar way but we concatenate the passage and corresponding document title as the input to T5 encoder. The query tower and the passage tower share parameters. The query to passage relevance is computed by the cosine similarity of their encodings. We optimize the model using the in-batch sampled softmax loss \cite{in-batch-neg}. The structure is shown in Figure~\ref{fig:de}. 

To benefit from both sparse model and dense neural model, we create the hybrid model by combining the encodings from two models in a principled way:
%\begin{small}
\begin{equation}
\begin{split}
    \nonumber
    \text{sim}(\textbf{q}^\text{hyb}, \textbf{c}^\text{hyb}) &= \langle \textbf{q}^\text{hyb}, \textbf{c}^\text{hyb} \rangle \\ \nonumber
    &= \langle [\textbf{q}^\text{bm25},\lambda\textbf{q}^\text{de}], [\textbf{c}^\text{bm25}, \textbf{c}^\text{de}] \rangle \\ \nonumber
    &=  \langle \textbf{q}^\text{bm25}, \textbf{c}^\text{bm25} \rangle + \lambda \langle \textbf{q}^\text{de}, \textbf{c}^\text{de} \rangle
\end{split}
\end{equation}
%\end{small}
where $\textbf{q}^\text{hyb}$ and $\textbf{c}^\text{hyb}$ are the hybrid encodings that concatenate the BM25 ($\textbf{q}^\text{bm25}$/$\textbf{c}^\text{bm25}$) and the dual encoder encodings ($\textbf{q}^\text{de}$/$\textbf{c}^\text{de}$) described above; and $\lambda$ is an interpolation hyperparameter that trades-off the relative weight of BM25 versus the dual encoder models.

To generate the training data for reranking model, we apply the hybrid retriever to the queries in the training set and retrieve top-K passages. We then sample $N$ negatives from retrieved result. In result, $C(i)$ is a passage list of size $N+1$ with one positive and $N$ negatives.

%\section{Evaluation}
\section{Experimental Setup}
%\textit{
%\begin{itemize}
%    \item Make it clear that NQ is ONLY used for training qgen model and is not used elsewhere.
    %\item Make it clear that the hybrid mixture parameter was tuned ONLY on MSMarco data.  We are aware that tuning would improve performance on many of these datasets due to the nature of the tasks (for some tasks term-based maching is more important; for others, semantic matching is important).
    %\item We do not choose hyperparameters for any targeted domain other than MSMarco.
    %\item MSMarco results are zero-shot results, NOT supervised DE results.
%\end{itemize}
%}
%\subsubsection{Corpora}
We evaluate our proposed approach on two tasks: first is the MS MARCO passage ranking task to understand how our approach performs in supervised setting when labeled training data is available. Second is the zero-shot retrieval on BEIR where no labeled data is available in the target domains.

\subsection{MS MARCO Passage Ranking}
\label{sec:msmarco_eval}
The MS MARCO passage ranking task aims to retrieve passages from a collection of web documents containing about 8.8 million passages. All questions in
this dataset are sampled from real and anonymized
Bing queries \cite{msmarco}. The dataset contains 532,761 and 6980 examples in the training and development set respectively. We report our results using MRR@10 metric on the
development set. We use the GTR-Large from \newcite{t5xr} as the dual encoder used in the hybrid retriever.

\subsection{Zero-shot Retrieval}
We also perform evaluation on the BEIR corpus \cite{beir}, a benchmark for zero-shot evaluation, to understand how our approach generalizes to out-of-domain setting. BEIR contains 18 evaluation datasets across 9 domains and no training data is available for those datasets. 
%\jing{describe the difference in zero-shot evaluation: apply msmarco reranker vs apply a zero-shot reranker (ours) }

To train the dual encoder used for hybrid retriever, we follow Ma et al. \shortcite{QGen} to generate synthetic training data from a query generator and extend with iterative training. Specifically, following \newcite{Promptagator}, we pretrain the dual encoder on C4 dataset \cite{2020t5} with the independent cropping task \cite{contriever}. We then fine-tune the dual encoder using synthetically generated queries. We apply a T5-based query generation model (QGen) on the passages in the target corpus to generate (synthetic query, passage) pairs. For each dataset, we iterate over every passage and treat every sentence as the target to generate synthetic queries. For large datasets, such as BioASQ and Climate-fever, we randomly sample 2 million passages for query generation. The statistics of BEIR and the number of synthetic data can be found in the Table \ref{tab:stats}.  We then use these data to train a dual encoder model $DE_{0}$.  
To filter low quality questions, we apply round-trip consistency for retrieval\cite{roundtrip, paq, Promptagator}. Specifically, given a synthetic query in the training data, we run 1-nearest neighbor search based on scores between the query and all passages using $DE_{0}$.  
If the neighbor is the one from which the query is generated, we keep that (query, passage).  
Otherwise, the (query, passage) pair is filtered. With the filtered data, we continue fine tune $DE_{0}$ to get the final dual encoder model $DE_{1}$. 

The QGen model is created by fine-tuning a general T5 model using question and passage pairs from Natural Question (NQ) \cite{nq}. This is the only place that we use supervised data in the pipeline. Particularly, we form the input of the encoder as ``Generate question \verb_>>>_ {title}.{passage} \verb_>>>_ {target sentence}'', and the output of the decoder is the corresponding question.  Here ``target sentence'' is the sentence that contains the short answer span, and ``passage'' corresponds to long answer and the passage of NQ. At inference time, given a passage, we iterate over every sentence as the target to generate diverse queries. Similar to PAQ \shortcite{paq}, we perform targeted generation, where knowing the location of the answer in a passage is important.

\begin{table}[]
\footnotesize
\resizebox{\linewidth}{!}{%
\begin{tabular}{r|c|cccc}
Datasets       & Domain      & \begin{tabular}[c]{@{}c@{}}\# Test \\ Query\end{tabular} & \begin{tabular}[c]{@{}c@{}}\# Corpus \\ Doc.\end{tabular} & \begin{tabular}[c]{@{}c@{}}\# Synth. \\ Ques. \end{tabular}  &  \begin{tabular}[c]{@{}c@{}}\# Synth. \\ Ques. \\ After \\ Filter \end{tabular} \\\hline
NQ             & Wiki   & 3452          & 2.68M              & 2.62M          &   976K \\
MS MARCO       & Misc.       & 6980          & 8.84M              & 3.74M     &   1.08M      \\
Trec-Covid     & BioMed & 50            & 171K               & 753K           &   352K \\
BioASQ         & BioMed & 500           & 14.91M             & 10.75M         &   3.44M \\
NFCorpus       & BioMed & 323           & 3.6K               & 22K            &   13K \\
HotpotQA       & Wiki   & 7405          & 5.23M              & 4.78M          &   3.19M \\
FiQA-2018      & Finance     & 648           & 57K                & 255K           &    157K \\
Signal-1M      & Twitter     & 97            & 2.86M              & 2.58M          &    1.03M \\
Trec-News      & News        & 57            & 595K               & 1.64M          &    618K \\
Robust04       & News        & 249           & 528K               & 2.65M          &    1.05M \\
ArguAna        & Misc.       & 1406          & 8.67K              & 41K          &    21K \\
Touché-2020    & Misc.       & 49            & 382K               & 1.34M          &    348K \\
Quora          & Quora       & 10000         & 523K               & 552K           &   400K \\
DBPedia-entity & Wiki   & 400           & 4.63M              & 4.76M         &    3.25M \\
SCIDOCS        & Science   & 1000          & 25K                & 136K       &    114K      \\
Fever          & Wiki   & 6666          & 5.42M              & 6.41M         &   4.22M  \\
Climate-Fever  & Wiki   & 1535          & 5.42M              & 6.41M         &   4.23M  \\
SciFact        & Science   & 300           & 5K                 & 27.3K      &   20K     \\
CQADupStack    & StackEx.    & 13145         & 457K               &  1.55M    &    1.18M          
\end{tabular}
}
\caption{The statistics of BEIR benchmark and number of synthetic questions before and after filtering.}
\label{tab:stats}
\end{table}

Results are obtained using the official TREC evaluation tool\footnote{https://github.com/usnistgov/trec\_eval}. We report normalised cumulative discount gain (nDCG@10) and recall@100 for all datasets.

%TODO check monoT5 performance: the bm25 results don't match

\subsection{Implementation Details}

The query generation model is initialized from T5 Version 1.1 XL model and we fine-tune using question and passage pairs from NQ \cite{nq} for 20,000 steps with a learning rate of 0.01, batch size 256 and dropout rate 0.1.

The BM25 model in our hybrid retriever is a unigram model. We use the WordPiece tokenizer and vocabulary from uncased BERT\textsubscript{base} of size 30522. We use K=0.9 and b=0.8 on all datasets. We only index the passages. The dual encoder model is initialized from the public pre-trained T5 Version 1.1 large model and pre-train on C4 for 200k steps with batch size 4096. We encode queries and passages into vectors of size 64 and 512 for all datasets except Climate-Fever and ArguAna. We use query length 100 for Climate-Fever and 512 for ArguAna. We use batch size of 5120 and train $DE_{0}$ for 100 epochs. $DE_{1}$ is initialized from $DE_{0}$ and we train 10 epochs. For the hybrid retrieval model, we use $\lambda = 600$ in searching from 50 to 750 with step size 50. 

The reranking models are initialized from T5 Version 1.1 models: Base and Large. Since we only use the encoders, the number of parameters are about 125M for Base model and 400M for Large model. We sample 50 negative examples from top 250 retrieved passage for MS MARCO and top retrieved passage from rank 10 to rank 210 for BEIR. 
%Due to the memory limitation, in each training iteration, we randomly pick 32 from 50 negative examples to append to the positive example. We try negative list size [16, 32, 64], and find that 32 provides the best results.
We use input sequence length as 512 for all datasets except ArguAna, for which we use 1024. We train the models for 20000 steps with batch size 64. 

All of the above hyperparameters were tuned on the MS MARCO development set and then applied to all other datasets, allowing for a true zero-shot setting (not to mention that not every dataset has a development set). From preliminary experiments, we found that tuning parameters for each dataset may achieve improved results due to the nature of each dataset. For example, when tuning the weight $\lambda$ between BM25 and dual encoder model in the hybrid retriever, the value should be smaller if we observe more term overlapping between questions and passage collection, such as SciFact and BioASQ, to emphasize the BM25 model.  However, for the results in this paper, we have use the parameters which were selected based on the MS MARCO dataset.

We implement the neural models using T5X\footnote{https://github.com/google-research/t5x} and we also use RAX \cite{rax}, a learning-to-rank framework for implementing the ranking losses in reranking models. For training, it takes about 6 hours to train a dual encoder model and 6.5 hours to train a reranking model of T5-Large size using Cloud TPU-V3.

\begin{table}[]
\begin{tabular}{c|cc}
\toprule
                                             & Model size     & MRR@10                     \\ \hline
{BM25 Anserini} &                & 0.1874 \\ \hline
HLATR                                        & RoBERTa\textsubscript{Large} & 0.3680 \\
MiniLM                                       & Distilled BERT & 0.3901 \\
monoT5                                   & T5\textsubscript{3B}          & 0.3980 \\
RankT5-EncDec                                & T5\textsubscript{Large}       &     0.3986                       \\
RankT5 (Ours)                                & T5\textsubscript{Large} 1.1   &     0.4222                       \\
HYRR                                         & T5\textsubscript{Large} 1.1   &     \textbf{0.4235}                  \\ 

\bottomrule
\end{tabular}
\caption{Reranking performance on MS MARCO Dev set in MRR@10.}
\label{tab:msmarco_result}
\end{table}

\begin{table*}[t!]
\centering
\small
\resizebox{\linewidth}{!}{%
\begin{tabular}{r|c|ccc|cc|cccc}
\toprule
     & \textbf{Retriever}     & \multicolumn{9}{c}{\textbf{Reranker}}    \\ \hline
     & & \multicolumn{3}{c}{\textbf{Supervised Baselines}} & \multicolumn{2}{|c}{\textbf{Unsupervised Baselines}} & \multicolumn{4}{|c}{\textbf{Ours}} \\ %\hline %\cline{2-10}
      & BM25 &  MiniLM  &  NQRR & MSHYRR &  SGPT &  UPR      &  HYRR  &  HYRR & HYRR\textsubscript{ft} &   HYRR\textsubscript{ft} \\ 
\textbf{Rerank Top K}   & (Anserini)  &  100   &  100   &  100   &  100 & 1000  &  100  &  100   &  100  &  1000      \\
%\textbf{Rerank Training Data} &   & MSMARCO   &     &     &    &  &  & NQ      & NQ      \\
\textbf{Model Size}     &  & 22M & 400M & 400M     & 6.1B      & 3B   & 125M & 400M  & 400M  & 400M \\ \hline
NQ    & 0.329   & 0.533     &  0.624\textsuperscript{\ddag}  & 0.569  & 0.401     & 0.454      & 0.532 & 0.555	 & 0.573	& \textbf{0.626}  \\ \hline
MS MARCO    & 0.228   & 0.413\textsuperscript{\ddag}     & 0.330 & \textbf{0.435\textsuperscript{\ddag}}  & 0.290     & 0.302     & 0.307 & 0.309	 & 0.319	& 0.344 \\
Trec-Covid        & 0.656   & 0.757     & 0.787  &  0.798   & 0.791     & 0.688  &  0.796 & 0.820	 & \textbf{0.822}	& 0.817  \\
BioASQ      & 0.465   & 0.523  &    0.507   & 0.554 & 0.547   & -  &  0.551 & 0.549	 & 0.554	&  \textbf{0.565} \\
NFCorpus    & 0.325   & 0.350     & 0.345 &  0.371   & 0.347     & 0.348  & 0.379  & 0.382 & 0.391	& \textbf{0.396} \\

HotpotQA    & 0.603   & 0.707     & 0.665 &  0.717   & 0.699     & \textbf{0.733}   & 0.706 & 0.707	 & 0.708	&  0.730 \\
FiQA-2018   & 0.236   & 0.347     & 0.397 & 0.411  & 0.401     & 0.444   & 0.408 & 0.437	 & 0.437	& \textbf{0.470}  \\
Signal-1M   & 0.330   & \textbf{0.338}     &   0.271 & 0.264  & 0.323     &  -   & 0.307 & 0.318	 & 0.304	&  0.280 \\
Trec-News   & 0.398   & 0.431     &   0.419 & 0.452  & \textbf{0.466}     &  -   & 0.437 & 0.453	 & 0.441	& 0.418  \\
Robust04    & 0.407   & 0.475     &   0.400 & 0.505 & 0.480     &  -  & 0.501 &	0.544 & 0.534	& \textbf{0.552}  \\
ArguAna     & \textbf{0.414}   & 0.311     &  0.107   & 0.351 & 0.286     & 0.372    & 0.344 & 0.342 &	0.382 & 0.326 \\
Touché-2020       & 0.367   & 0.271     &  0.363  & 0.467 & 0.234     & 0.206  &  0.368 & \textbf{0.384}	 & 0.366	&  0.287 \\
Quora       & 0.789   & 0.825     &  0.807 &  0.637   & 0.794     & 0.831    & 0.861  & 0.867 & \textbf{0.869}	& 0.868  \\
DBPedia-entity    & 0.313   & 0.409     &  0.391 & 0.402  & 0.370     & \textbf{0.534}  & 0.385 & 0.403	 & 0.408	&  0.443 \\
SCIDOCS     & 0.158   & 0.166     &  0.182 & 0.184  & 0.196     & 0.170  &  0.183 & 0.187 &	0.198 & \textbf{0.201}  \\
Fever       & 0.753   & 0.819     &  \textbf{0.871} & 0.825 & 0.725     & 0.591    & 0.868 & 0.861	 & 0.851	&  0.856 \\
Climate-Fever     & 0.213   & 0.253     & \textbf{0.321} &  0.262   & 0.161     & 0.117   & 0.272 & 0.294	 & 0.275	& 0.271  \\
SciFact     & 0.665   & 0.688     &   0.696 & 0.745  & 0.682     & 0.703   & 0.734 & 0.754 &	0.755 &  \textbf{0.767}  \\
CQADupStack       & 0.299   & 0.370     &  0.397  &  0.368  & 0.420     & 0.416   &  0.398  &	0.416 &	0.421 & \textbf{0.444} \\ \hline
Average &  0.418  &  0.473   & 0.467 & 0.490  & 0.453     &  0.461   & 0.491 &	0.504 &	0.506 & 0.508 \\ 
Average w/o NQ & 0.423  & 0.470    &  0.459 & 0.486   & 0.456  & 0.461  &  0.489 & 0.501	 & 0.502	& 0.502 \\ \hline
\multicolumn{2}{c|}{Avg. improvement on BM25} & 4.63\%  & 3.54\% & 6.26\% & 3.29\%  & 3.11\%  & 6.58\% & 7.81\%	 & 7.86\%	& 7.87\%    \\
\bottomrule
\end{tabular}
}
\caption{Reranking performance on BEIR in NDCG@10. \ddag~indicates the in-domain performances. The results of baseline models are copied verbatim from the original papers. ``-'' indicate results that are not reported. Baseline models rerank the top-100 passages from BM25 except UPR that reranks the top-1000 passages.}
\label{tab:reranking}
\end{table*}

\section{Results and Discussion}
\subsection{Results on MS MARCO}
\label{sec:msmarcoexperiments}

To understand the effectiveness of our proposed approach, we fix the first-stage retrieval system and compare the reranking performance. Table~\ref{tab:msmarco_result} shows the performance of our proposed reranker on reranking BM25 top-1000 results. The BM25 results in row 1 is obtained from Anserini \cite{anserini} toolkit\footnote{https://github.com/castorini/anserini} with parameters: k=0.82, b=0.68 following other baselines. The results of several strong baselines are shown in row 2-6. \textbf{HLATR} \cite{hlatr} extends the retrieval-and-rerank pipeline with an additional ranking module by using the features from retrieval and reranking stages. It achieves top performance on MS MARCO leaderboard. \textbf{MiniLM} \cite{minilm}, which is a cross-encoder reranking model 
%trained on supervised MS MARCO data and 
distilled from an ensemble of three teacher models: BERT-base, BERT-large and ALBERT-large. The other baselines are T5-based models: \textbf{monoT5} \cite{2020t5,nopara} and \textbf{RankT5-EncDec} \cite{rankT5} adopt the encoder-decoder architecture. Our model adopts RankT5's encoder-only variant as described in Section \ref{sec:reranking}. To compare with RankT5 model fairly, we implement our version of \textbf{RankT5 (ours)}, which shares the same architecture and parameter settings as \textbf{HYRR}. They only differ from the training data generation. Our reproduced RankT5 generates training data from dual encoder retriever. From row 7, we can see that HYRR improves over BM25 performance by 23.6\% in MRR@10, and outperforms other baselines. This demonstrates that our proposed training framework is effective in supervised setting.

\subsection{Results on BEIR}
Similar as evaluation on MS MARCO, we fix the first-stage retrieval system and compare the reranking performance. Table~\ref{tab:reranking} shows the reranking performance of our proposed reranker. The BM25 results in Col.1 are obtained from Anserini toolkit with parameters:k=0.9 and b=0.4 following other baselines.  

We consider several reranking models that can be categorized in two zero-shot settings as baselines. The first one is to train a supervised reranker and perform inference on the target domains directly, and the results are shown in Col. 2-4. We choose \textbf{MiniLM} \cite{minilm}, which reranks the top 100 retrieved passages from BM25. The result on MS MARCO is considered as in-domain for this model. We train a T5-based reranker \textbf{NQRR} using supervised NQ data and the model structure is the same as the one described in Section \ref{sec:reranking}. The result on the NQ dataset is considered in-domain for the NQRR.  We also use our reranker \textbf{MSHYRR} trained using our proposed approach on MS MARCO from Section \ref{sec:msmarco_eval} as baseline.

The second setting is unsupervised, and we choose two models to compare. \textbf{SGPT} \cite{sgpt} uses a pre-trained GPT model in a cross-encoder setting for reranking.  Specifically, query text and passage text are concatenated and fed into pre-trained GPT model. The log probability is used as the score for reranking. Col. 5 shows the results of reranking top 100 retrieved passages with model size of 6.1 billion parameters. \textbf{UPR} \cite{upr} also uses a pre-trained language model for reranking. UPR uses a T5 model by taking the passage and a simple prompt as input and uses the average likelihood of query tokens conditioned on the passage as the score for reranking. Col. 6 shows the results of reranking top 1000 retrieved passages using T0 model \cite{T0} with 3 billion parameters. 

\begin{table*}[]
\centering
\small
\begin{tabular}{r|cccc|cccc}
\toprule

Retriever  $\downarrow$  & No Reranker & BM25RR & DERR & HYRR & No Reranker & BM25RR & DERR & HYRR \\\hline
& \multicolumn{8}{c}{\textbf{MS MARCO}} \\ \hline
& \multicolumn{4}{c|}{\textbf{MRR@10}}         & \multicolumn{4}{c}{\textbf{nDCG@10}}                \\
BM25   &   0.187   &  0.375     &  0.422    &   \textbf{0.424}    & 0.234 	& 0.438	& 0.485	&  \textbf{0.486}  \\
DE     &  0.378    &  0.350     & \textbf{0.440}    &   \textbf{0.440}    & 0.445 	& 0.411 	& \textbf{0.507} & \textbf{0.507} \\
Hybrid &  0.390    &  0.351     &  0.438    &   \textbf{0.440}    & 0.457 	& 0.413	& 0.506	& \textbf{0.508} \\ \hline
& \multicolumn{8}{c}{\textbf{SciFact}} \\ \hline
& \multicolumn{4}{c|}{\textbf{nDCG@10}}         & \multicolumn{4}{c}{\textbf{Recall@100}}                \\

BM25   &  0.677     &   0.750      &   0.742    &    \textbf{0.752}   &  0.918	& 0.916	& 0.923	& \textbf{0.946}  \\
DE     &  0.597     &   \textbf{0.755}      &   0.745    &    0.752   &  0.903	& 0.919	& 0.923	& \textbf{0.946}  \\
Hybrid &  0.706     &    0.753     &   0.744    &    \textbf{0.759}   &  0.941	& 0.931	& 0.931	& \textbf{0.958} \\
\bottomrule
\end{tabular}
\caption{Ablation results on MS MARCO and SciFact.}
\label{tab:ablation}
\end{table*}

For our models, we show results of two training settings. \textbf{HYRR} is a fine-tuned vanilla T5 model, while \textbf{HYRR\textsubscript{ft}} is a fine-tuned \textbf{NQRR} model. The results are shown in Col. 7-10. As we can see, our proposed model outperform the baseline models in 10 of 18 out-of-domain datasets. The best setting is reranking the top 1000 retrieved passages using \textbf{HYRR\textsubscript{ft}} and the improvement of nDCG@10 over BM25 is from 2\% - 23.4\%, and in average 7.87\%. Comparing Col. 8 and Col. 9, we can see that \textbf{HYRR} already provides strong performance on many datasets, and fine-tuning an out-of-domain supervised reranker achieves an additional performance boost.  We note that the results on NQ cannot be considered completely out-of-domain even for our \textbf{HYRR} as the question generation model used to generate training data for the hybrid retriever is trained on the NQ dataset. Similarly, \textbf{UPR}'s results on HotpotQA cannot be considered completely out-of-domain since T0 uses HotpotQA for training. The evaluation on BEIR demonstrates the effectiveness of our proposed method in zero-shot settings. 

\begin{table*}[]
\centering
\small

\resizebox{\linewidth}{!}{

\begin{tabular}{l|p{16cm}}
\toprule

Query   & what mlb team play in \\ \hline

HYRR   &  \textbf{P1} major league baseball (\textbf{mlb}) is a professional baseball organization, ... a total of 30 teams now \textbf{play in} the american league (al) and national league (nl)...  \\
 &  \textbf{P2} in the united states and canada, professional major league baseball (\textbf{mlb}) teams are divided into the national league (nl) and american league (al), each with three divisions..." \\ \hline
BM25RR     &  \textbf{P3} major league baseball. the 16 teams were located in ten cities, all in the northeastern and midwestern united states: new york city had three teams and boston, chicago, ....  \\ 
 & \textbf{P4} list of current major league baseball stadiums. the newest \textbf{mlb} stadium is suntrust park in cumberland, georgia, home of the atlanta braves, which opened for the 2017 season. \\ \hline
DERR & \textbf{P5} major league baseball. the arizona diamondbacks, ..., are a major league baseball (\textbf{mlb}) franchise that play in the west division....  \\ 
 & \textbf{P6} boston red sox. the yankees compete in major league baseball (\textbf{mlb}) as a member club of the american league (al) east division.... \\ 
\bottomrule
\end{tabular}
}
\caption{Reranker predictions on MS MARCO}
\label{tab:example}
\end{table*}

\subsection{Ablation}

To show the robustness of HYRR, we conduct an ablation experiment. We train rerankers using the training data generated from the BM25 or the dual encoder model, namely \textbf{BM25RR} and \textbf{DERR}. Those two variants are commonly seen in many pipelined retrieval systems, where rerankers are simply trained upon the first-stage retriever. We train them using the same training setting for HYRR and then apply them on three retrievers: the BM25 model, the dual encoder model (DE) and the hybrid retriever, respectively. We experiment on both supervised setting and zero-shot setting. The results on MS MARCO are shown in the top section of Table~\ref{tab:ablation}. As we can see HYRR provides the most performance gain over all three retrievers on both MRR@10 and nDCG@10. The BM25RR improves the performance on BM25 while hurts the other two. DERR achieves best performance when we apply it to DE. It also improves the other two retrievers but not as much as HYRR. This shows that HYRR not only outperforms the other two rerankers but also is effective on different retrievers.

\begin{table*}[]
\small
\centering
\resizebox{\linewidth}{!}{%
\begin{tabular}{l|p{1.6cm}|p{6cm}|p{6cm}}
\toprule
\textbf{Corpus}    &  \textbf{Query Type}  &  \textbf{Test Queries}     & \textbf{Synthetic Questions}                   \\ \hline

FiQA-2018 & Financial Question & How does unemployment insurance work? & what is the advantage of stocks over bonds \\ \hline

BioASQ & Biomedical Question & List types of DNA lesions caused by UV light. &  how many subjects in the silent lacunar lesion study \\ 
& & Does deletion of cohesin change gene expression? & what happens to amino acids in the hypothermic kidney \\ \hline 

Fever & Claim & The Hunger Games are not based on a novel. & when did bag raiders album come out in australia \\ 
& & Pharmacology is a science. & when did angela lien win her first olympic medal \\ \hline 

SCIDOCS & Paper Title & Multi-task Domain Adaptation for Sequence Tagging &  what is the use of blockchain in software architecture \\ 
 & & HF outphasing transmitter using class-E power amplifiers & what is the relationship between test anxiety and mathematics performance \\ \hline \hline
 
 \rule{-2pt}{10pt}

Signal-1M & \begin{tabular}[c]{@{}c@{}} News \\ Headline \end{tabular}  & Inside Trevor Noah’s final test run for the new ‘Daily Show’ & how many food items at the state fair \\
%& & Your Money Adviser: For College Students, Choosing a Bank Account Can Be a Minefield &  what is the right hair color for a girl \\
& & Fancy Cooking & where is book week held in the uk \\ \hline

Touché-2020 & Controversial Questions  & Is golf a sport? & who is the democrat in the 2020 presidential debate  \\
& & Is drinking milk healthy for humans? & what kind of music can you debate on wikipedia  \\ \hline

Arguana   & Argument        & Small businesses need advertisements to make their products known. If there wasn't advertising then small businesses would have no chance at all to make their product well known. Adverts can actually level the playing field - if you have a good new product, and market it in a clever way then it doesn't matter how small your company is, you can still make consumers interested.... & who benefits from the mass media and advertising   \\ 

\bottomrule
%The more you restrict the freedom of information, the more this helps the large companies who everyone already knows about.

\end{tabular}
}
\caption{Examples of test questions in some corpora and generated synthetic questions.}
\label{tab:qgen}
\end{table*}

When examining the ranking outputs on MS MARCO, we find that top ranked predictions from HYRR is more semantically and lexical relevant to the query, while BM25RR and DERR sometimes return partially related predictions. For example, query ``what mlb team play in'' which asks for general information about major league baseball(mlb) shown in Table~\ref{tab:example}, HYRR returns the correct predictions, while DERR returns predictions about individual team in mlb, and BM25RR returns less relevant information about mlb. 

We pick SciFact from BEIR as an example for zero-shot setting. The results are shown in bottom part of Table~\ref{tab:ablation}, and we observe the similar trends on other datasets in BEIR. Similarly, HYRR improves both nDCG@10 and Recall@100 over all three retrievers. Specifically, it outperforms BM25RR and DERR not only when we apply it to the matched Hybrid retriever but also the other two retrievers. This believe is the evidence that the robustness of the training data for the reranker is carried over to the robustness of the reranker itself. 

In addition, to understand the benefit to use hybrid retriever for training data generation, we conduct another ablation experiment. We mix the training data generated from the BM25 and the dual encoder model in 1:1 ratio and train a reranker. We evaluate on MS MARCO and the model achieves 0.417 in MRR@10 and 0.480 in nDCG@10 when reranking BM25 top 1000. When comparing with the results in row 1 from Table~\ref{tab:ablation}, we can see that our approach significantly outperforms the approach that simply applying training data ensemble. 

%\jing{another ablation experiment is to mix negatives from BM25 and DE and train a reranker, in order to show that hybrid retriever provide a different distribution of negatives}

\subsection{Discussion}

\textbf{Robustness against Initial Checkpoints.} While when reranking the BM25 outputs in the zero-shot settings, our best reranking model is the fine-tuned model on a supervised reranking model (\textbf{NQRR}), it is worth to note that the zero-shot reranking model also achieves strong performance and outperforms \textbf{NQRR} on many datasets as shown in Table~\ref{tab:reranking}. 
%The difference between \textbf{HYRR} and \textbf{HYRR\textsubscript{ft}} is even smaller when we evaluate the end-to-end performance as shown in Table~\ref{tab:e2e}. 
This shows that our proposed training framework is useful in many practical situation when supervised data is not available.

\textbf{Robustness against Model Sizes.} Regarding the model size, recent work shows that increasing model size results in large gains in zero-shot performance for reranking models. A 10\% improvement in average nDCG@10 over all datasets in BEIR can be seen when the model size increases from T5-small to T5-3B \cite{nopara}. From Table~\ref{tab:msmarco_result} and \ref{tab:reranking}, we can see that our proposed model is smaller, but more effective than some larger models.

\textbf{Robustness against Retrievers.} We also show that despite the common wisdom that the best reranking model will be one trained on the same retriever as will be used at inference time, a model trained with hybrid retrieval output is strong for both BM25 and dual encoder retrievers.  Reranking BM25 resulted in superior performance despite the mismatch between the retriever used for generating training data and that used for retrieval. The ablation experiment also confirms this observation.

%\subsection{Error Analysis}

\textbf{Robustness against Query Generation.} In addition, we analyze the synthetically generated questions and show that our proposed model generalizes to different domains and different query types. In Table~\ref{tab:qgen}, we show examples of queries from the test set and the synthetically generated questions. As we can see, when synthetic questions are similar to test queries, our model performs very well even though the target domain is very different from the domain where the query generator is trained. For example, questions in FiQA-2018 are from investment topic and questions in BioASQ are scientific questions in biomedical domain. Both retriever and reranker performs very well on those two datasets. Interestingly, we find that our model also performs well in some cases where test queries and synthetic questions are of different type. For example, Fever asks to retrieve passages to verify given claims. The test queries are claims rather than questions. A similar pattern exists for SCIDOCS where the queries are paper titles.  Our retrieval and ranking models learn the relevance between queries and passages despite the test queries not being questions.

However, when the synthetic question is very different from the test queries, our model fails.  For example, Signal-1M asks to retrieve relevant tweets given news headlines.  Another example is Touché-2020, where the task is to retrieve relevant arguments given a question on a controversial topic. Also in Arguana, the task is to retrieve counterargument to an argument, and queries are long text passages. The nuance of the task itself and whether it requires term-based or semantic abstractions to match queries to passages appears to be where our QGen approach is ineffective.

% \textit{
% Need to turn the following into an actual discussion section.
% \begin{itemize}
%     \item While the best system is the fine-tuned NQ model, the pure zero-shot model performs very well with only a marginal gain by including the pretrained NQ model.
%     \item Reranking BM25 resulted in superior results despite the mismatch between the retriever used for generating training data and that used for retrieval.  Depending on results, focus on this robustness from the SciFact results.
%     \item The smaller t5 models we use can beat much larger general models.
    
% \end{itemize}
% }

\section{Conclusion}
We proposed a generic training framework for rerankers based on a hybrid retriever. While the hybrid retriever is composed of term-based and neural models, the reranker is a neural cross-attention model which learns from negatives examples generated by the hybrid retriever.  
%We train both the neural retriever and reranker in a zero-shot setting, where a synthetic query generator is used to create positive examples in each target domain. 
The proposed approach is robust and outperforms several strong baselines on MS MARCO passage ranking task and BEIR benchmark dataset, which demonstrates that it is practical and generalized.
We observe that a model trained with robust training instances (in this case, from the hybrid retriever) produces a reranker that outperforms matched-training rerankers for term-based or neural retrievers. 

\bibliography{anthology,custom,zsrr}
\bibliographystyle{acl_natbib}

\appendix

\section{Loss function for T5 dual encoder}

We train the duel encoder model using an in-batch sampled softmax loss:
\begin{equation}
\footnotesize
  \ell = \frac{e^{\text{sim}(q_i, \hat{p}_i)/ \tau}}{\sum_{j \in \mathcal{B}} { e^{\text{sim}(q_i, \hat{p}_j) / \tau} } } \nonumber
  \label{eq::loss}
\end{equation}
where 
% the similarity scoring function \textit{sim} is the cosine similarity between the embeddings of $q_i$ and $p_i^{+}$.
$q_i$ is the question encoding and $\hat{p}_i$ is the encoding of gold passage.
\textit{sim} is cosine similarity,
$\mathcal{B}$ is a mini-batch of examples, and $\tau$ is the softmax temperature.

\end{document}